%% file: main.tex
\pdfoutput=1 %

\documentclass[letterpaper, 10 pt, conference]{ieeeconf}
\IEEEoverridecommandlockouts
\usepackage{graphicx}
\usepackage{amsmath,amssymb}
\usepackage{booktabs}
\usepackage[T1]{fontenc} 
\usepackage[colorlinks=true,citecolor=blue,linkcolor=blue]{hyperref}
\hypersetup{pdftitle={Déjà-Events: Multi-view 3D Geometry with Event Cameras via Sparsity-Aware Adaptive Gating},
  pdfauthor={Arda Cinar Demirtas, Pablo Varela Bernal, Marc Pollefeys, Sunghwan Hong}}
\usepackage{xcolor}
\usepackage{tikz}
\usepackage{standalone}
\usetikzlibrary{arrows.meta,positioning,fit,backgrounds,calc,shapes.geometric}

\definecolor{stagecol}{RGB}{235,241,250}
\definecolor{loopcol}{RGB}{255,244,224}
\definecolor{headcol}{RGB}{233,245,233}
\definecolor{addcol}{RGB}{252,228,236}
\definecolor{addborder}{RGB}{194,24,91}
\definecolor{gatecol}{RGB}{255,224,178}

\tikzset{
  block/.style={draw, rounded corners=2pt, fill=white, minimum height=8mm, minimum width=26mm, align=center, font=\small},
  add/.style={draw, circle, fill=white, minimum size=4mm, inner sep=0pt, font=\scriptsize},
  gate/.style={draw, circle, fill=gatecol, minimum size=4mm, inner sep=0pt, font=\scriptsize},
  resline/.style={-{Stealth[length=2mm]}, thick},
  branchline/.style={-{Stealth[length=2mm]}, thin},
}
\usetikzlibrary{arrows.meta,positioning,fit,backgrounds,calc}

\newcommand{\tabnote}[1]{\par\smallskip\noindent\parbox{\linewidth}{\scriptsize #1}}

\makeatletter
\long\def\@makecaption#1#2{%
\ifx\@captype\@IEEEtablestring%
{\centering\footnotesize #1: {\scshape #2}\par}%
\@IEEEtablecaptionsepspace%
\else
\@IEEEfigurecaptionsepspace%
\setbox\@tempboxa\hbox{\footnotesize #1.~~ #2}%
\ifdim \wd\@tempboxa >\hsize%
\setbox\@tempboxa\hbox{\footnotesize #1.~~ }%
\parbox[t]{\hsize}{\footnotesize \noindent\unhbox\@tempboxa#2}%
\else%
\ifcenterfigcaptions \hbox to\hsize{\footnotesize\hfil\box\@tempboxa\hfil}%
\else \hbox to\hsize{\footnotesize\box\@tempboxa\hfil}%
\fi\fi\fi}
\makeatother

\title{\LARGE \bf
  Déjà-Events: Multi-view 3D Geometry with Event Cameras via Sparsity-Aware Adaptive Gating
}

\author{Arda Cinar Demirtas$^{1}$, Pablo Varela Bernal$^{2}$, Marc Pollefeys$^{1}$ and Sunghwan Hong$^{1}$\\
  {\normalsize $^{1}$ETH Zürich \qquad $^{2}$TU Delft}}

\begin{document}

\maketitle
\thispagestyle{empty}
\pagestyle{empty}

\begin{abstract}
Event cameras keep capturing structure where conventional cameras blur or saturate, which makes them attractive for 3D perception in fast or poorly lit scenes. Event datasets with dense 3D ground truth, however, are far smaller and less diverse than the image collections behind feed-forward 3D reconstruction models. We present Déjà-Events, an event-only model that predicts depth, camera pose and pointmaps from multiple views in a single forward pass. Rather than relying on scarce 3D labels, we distil it from Déjà View, a feed-forward RGB model that refines its tokens by repeatedly applying one shared transformer block. On paired RGB and event recordings, the event student learns to reproduce the frozen teacher's final tokens, which requires no depth or pose ground truth and trains only about 1\% of its parameters. Because events are dense at edges and moving texture but nearly absent in textureless regions, we further introduce sparsity-aware adaptive gating, which adapts the shared block's gates to events and adjusts them per patch by event density, so that accuracy holds as refinement steps are added. Trained only on simulated data, Déjà-Events achieves the lowest pointmap error among the event-based methods we evaluate, on EventScape and zero-shot on the real MVSEC and DSEC datasets. On DSEC it halves the error of the strongest baseline (0.15 vs.\ 0.30 AbsRel) with about a tenth of its parameters. It also has the lowest depth error on EventScape and DSEC and the lowest trajectory error on EventScape.
\end{abstract}

\input{sections/intro}
\input{sections/related}
\input{sections/method}
\input{sections/experiments}
\input{sections/results}
\input{sections/discussion}
\input{sections/conclusion}

\bibliographystyle{IEEEtran}
\bibliography{refs}

\end{document}

%% file: sections/intro.tex
\section{Introduction}
\label{sec:intro}

Robots that move fast or operate in poor lighting need the 3D geometry of their surroundings, and their own motion through it, with little delay, yet these are exactly the conditions in which frame cameras blur, saturate or under-expose~\cite{bartolomei2025depthanyevent,huang2026event3r}. Event cameras avoid much of this: each pixel asynchronously reports brightness changes as they happen, giving microsecond temporal resolution and a much higher dynamic range than conventional cameras~\cite{gallego2022eventsurvey,hidalgocarrio2020e2depth}. These make events an attractive modality for 3D reconstruction in such scenes~\cite{ren2026eventvggt}.

\begin{figure}[t]
  \centering
  \includegraphics[width=\linewidth]{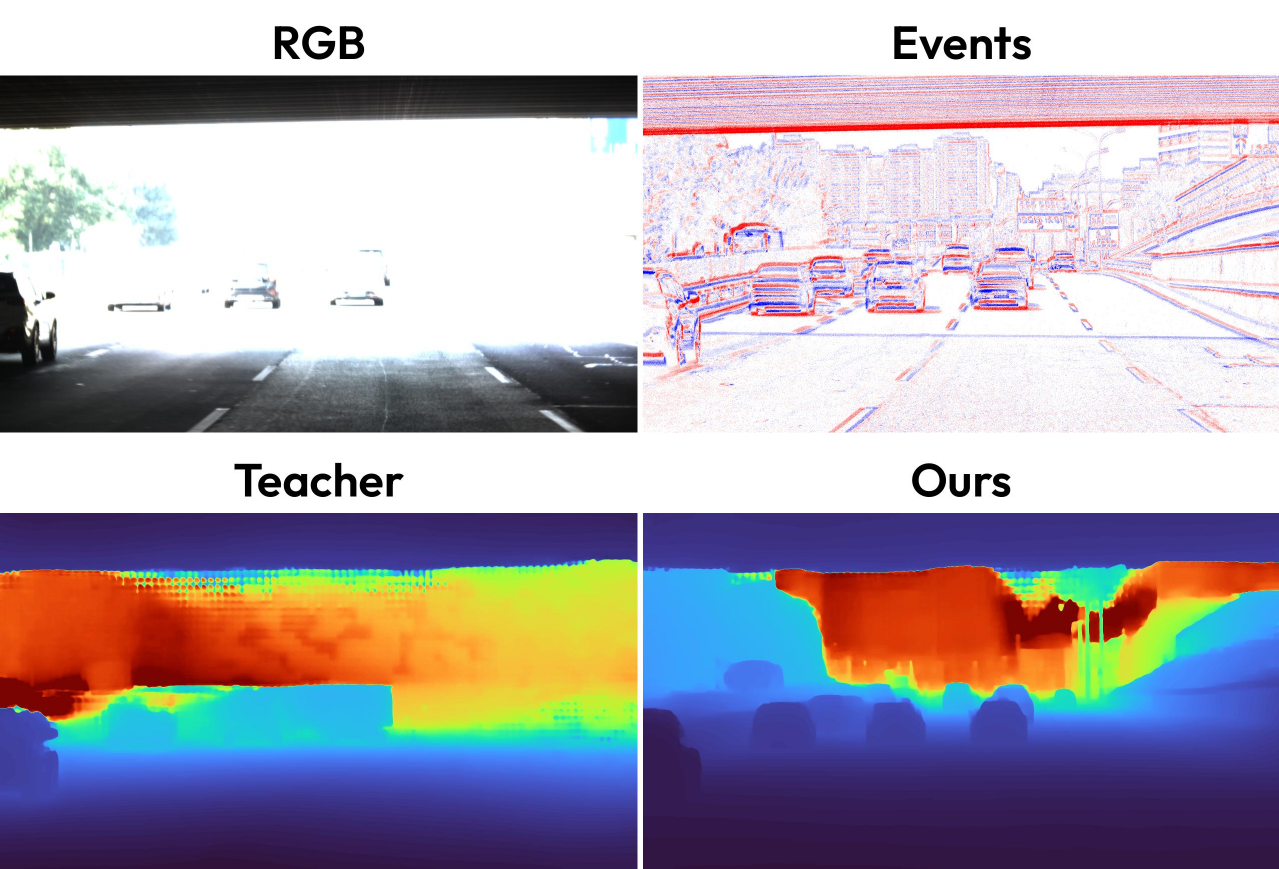}
  \caption{\textbf{Qualitative results on PEOD}~\cite{cui2026peod}. At a tunnel exit the RGB frame is saturated
  across most of the road and the vehicles ahead, while the event stream
  retains structure. In the blown-out area the RGB teacher predicts a
  flat, ground-plane-like depth, while the event student keeps the
  vehicles and the tunnel structure. Neither model was trained on PEOD.}
  \label{fig:teaser}
\end{figure}

Feed-forward reconstruction models predict camera poses, depth and 3D pointmaps from a set of images in a single forward pass~\cite{wang2025vggt,keetha2025mapanything,burzio2026dejaview}. They are trained on large and diverse image collections with 3D annotations~\cite{wang2025vggt,burzio2026dejaview}, whereas event datasets with dense 3D ground truth are much smaller and less diverse~\cite{bartolomei2025depthanyevent,zhu2025eventdam,ren2026eventvggt}. We therefore distil geometry from an RGB model instead: a frozen RGB teacher sees RGB frames, and an event student learns to reproduce what the teacher computes from events paired with those frames alone, transferring the teacher's priors to events~\cite{bartolomei2025depthanyevent,ren2026eventvggt}. Because the student inherits those priors but not the sensor, it can keep predicting structure where the teacher's frames saturate (Fig.~\ref{fig:teaser}).

Earlier methods that distil geometry into events used monocular depth models~\cite{bartolomei2025depthanyevent,zhu2025eventdam} or VGGT~\cite{ren2026eventvggt} as the teacher. Ours is Déjà View~\cite{burzio2026dejaview} (DVLT), which refines the patch tokens of all images by reusing one shared transformer block $K$ times, matching much larger feed-forward models with a fraction of their parameters. Because $K$ is set at inference time, the same model can also trade accuracy for compute~\cite{burzio2026dejaview}.

In a well-exposed RGB frame almost every patch carries information, whereas events fire only where brightness changes, mostly at edges and moving texture, leaving textureless areas nearly empty~\cite{hidalgocarrio2020e2depth,bartolomei2025depthanyevent,zhu2025eventdam}. In DVLT, each step's update is scaled by a gate that depends only on the step and is thus identical for every patch. We expect this to suit events poorly, as sparse patches may contribute little early in the refinement and benefit from context once nearby dense regions are resolved. We therefore introduce sparsity-aware adaptive gating (SAAG), which fine-tunes this gate for events and adds a small per-patch correction conditioned on event density and the current step.

In summary, our contributions are as follows.
\begin{itemize}
  \item Déjà-Events, an event-only feed-forward geometry model distilled from DVLT that predicts depth, camera pose and pointmaps in a single forward pass, without depth or pose ground truth while training about 1\% of its parameters.
  \item State-of-the-art event-only 3D reconstruction: of the event-based methods we evaluate, Déjà-Events has the lowest pointmap error on EventScape and, zero-shot, on the real MVSEC and DSEC datasets, with about a tenth of EventVGGT's parameters.
  \item Sparsity-aware adaptive gating, which conditions the gates of the shared block on the event density of each patch, so that the recursion adapts to where events actually fire.
\end{itemize}

%% file: sections/related.tex
\section{Related Work}
\label{sec:related}

\paragraph{Feed-forward 3D reconstruction} Feed-forward reconstruction models regress cameras and 3D structure directly from images. DUSt3R~\cite{wang2023dust3r} and MASt3R~\cite{leroy2024mast3r} predict pointmaps for pairs of images, and DUSt3R fuses larger sets through a global alignment optimisation, whereas VGGT~\cite{wang2025vggt}, $\pi^3$~\cite{pi3_2025}, MapAnything~\cite{keetha2025mapanything} and Depth Anything 3~\cite{lin2025depthanything3} process all views jointly and recover cameras, depth and points in one forward pass through a deep stack of transformer layers. DVLT~\cite{burzio2026dejaview} instead applies a single shared block for a number of steps chosen at inference.

\paragraph{Geometry from events} Early event-based methods recovered structure geometrically, for example by a space sweep along a known camera trajectory in EMVS~\cite{rebecq2018emvs}. Learned depth estimators either reconstruct intensity images from events with E2VID~\cite{rebecq2019e2vid} and apply a frame-based depth network to them~\cite{hidalgocarrio2020e2depth}, or regress depth directly, as E2Depth~\cite{hidalgocarrio2020e2depth} does with a recurrent convolutional network and EReFormer~\cite{liu2022ereformer} with a recurrent vision transformer. Their depth maps can be combined with camera poses from event visual odometry such as DEVO~\cite{klenk2024devo} to form pointmaps. Other models predict pointmaps directly by adding events to the feed-forward models above, while keeping RGB as an input: EAG3R~\cite{wu2025eag3r} adds events to handle dynamic and low-light scenes, and Interp3R~\cite{guo2026interp3r} uses them to interpolate a frame-based model's pointmaps between frames. A concurrent work, Event3R~\cite{huang2026event3r} instead reconstructs from events alone, training a pairwise backbone initialised from DUSt3R with ground-truth and self-supervised losses, and fusing more than two views by global optimisation. EvGGS~\cite{wang2024evggs} also works from events alone, but needs known camera poses to place the 3D Gaussians it predicts.

\paragraph{Distillation from RGB foundation models} Another line of work, motivated partly by the scarcity of event recordings with dense 3D labels, distils RGB foundation models which requires only paired frames and events. Depth AnyEvent~\cite{bartolomei2025depthanyevent} supervises an event student with the depth that a monocular RGB model predicts on aligned frames. EventDAM~\cite{zhu2025eventdam} also distils a monocular model; it scores the event sparsity of each patch and uses this score to mix RGB and event features into auxiliary depth targets, to weight feature distillation, and to enforce consistency across sparsity levels. EventVGGT~\cite{ren2026eventvggt} distils VGGT, aligning the features of an event student with the teacher's alongside a depth loss on randomly mixed RGB and event features and a temporal-consistency loss.

\paragraph{Input-adaptive computation} Event-specific networks exploit sparsity for efficiency by recomputing activations only where new events arrive, as AsyNet~\cite{messikommer2020asynet} does with sparse convolutions and AEGNN~\cite{schaefer2022aegnn} on an event graph. Adaptive computation methods instead learn how much computation to spend: Adaptive Computation Time~\cite{graves2016act} and PonderNet~\cite{banino2021pondernet} learn how many recurrent steps to take, Universal Transformers~\cite{dehghani2018universal} apply one shared block recurrently under such a halting rule, and Mixture-of-Depths~\cite{raposo2024mixtureofdepths} learns which tokens may skip a layer. Among the distilled models above, only EventDAM~\cite{zhu2025eventdam} makes event sparsity explicit, and it does so in the training objective; REALM~\cite{polizzi2026realm} and EventVGGT~\cite{ren2026eventvggt} align features without conditioning on it. We instead let sparsity steer the refinement itself, by conditioning the shared block's gates on the event density of each patch, which adds no loss term and only a few thousand parameters.

%% file: sections/method.tex
\section{Method}
\label{sec:method}

\begin{figure*}[t]
  \centering
  \includegraphics[width=\textwidth]{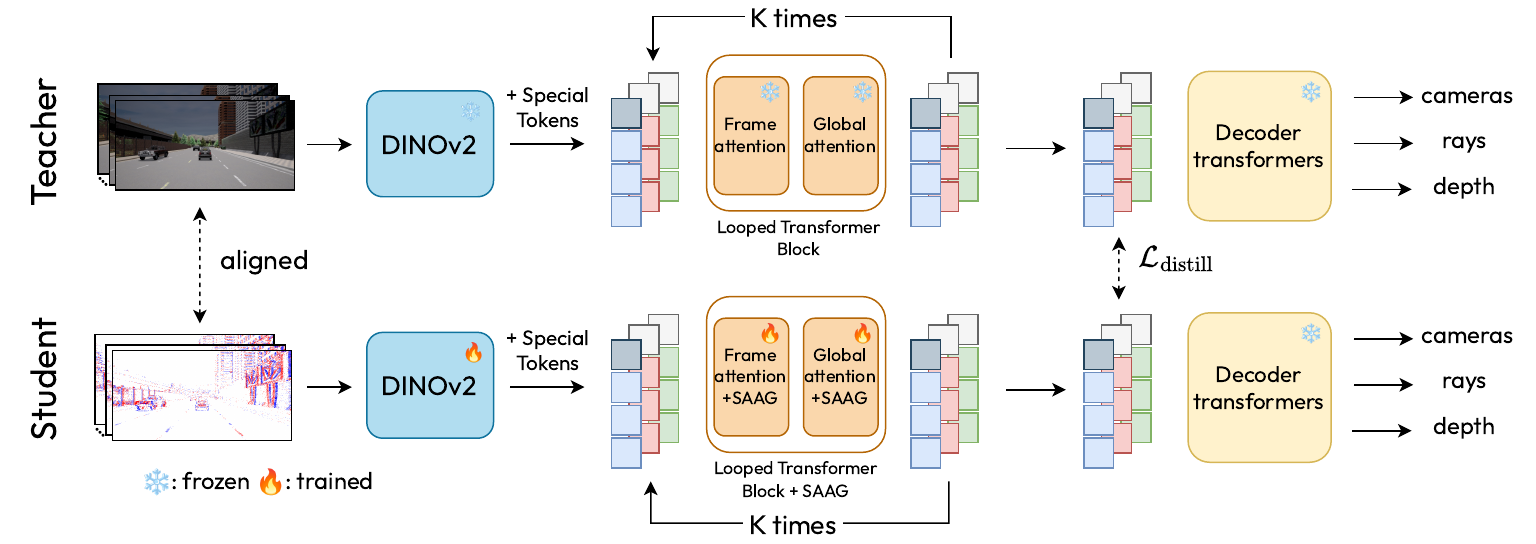}
  \caption{Overview. A frozen RGB DVLT teacher (top) and an event student with
  the same architecture (bottom) process aligned RGB frames and events. In each
  training iteration both run the shared block for the same sampled number of
  steps $K$, and the loss matches their final tokens
  (Sec.~\ref{sec:method-distill}). The student trains low-rank adapters in
  DINOv2 and in the shared block, and its gates are unfrozen and corrected per
  patch in both attention blocks
  (Sec.~\ref{sec:method-patchwise}).}
  \label{fig:architecture-overview}
\end{figure*}

\subsection{DVLT and its adaptation to events}
\label{sec:method-dvlt}

Fig.~\ref{fig:architecture-overview} gives an overview. DVLT~\cite{burzio2026dejaview} embeds the patches of $S$ views with DINOv2~\cite{oquab2023dinov2}, adds a camera token~\cite{wang2025vggt} and four register tokens~\cite{darcet2024registers} per view, and refines these $C$-channel tokens by applying one shared block $F$ for $K$ steps,
\begin{equation}
  z_k = F\big(z_{k-1};\, t_k, t_{k+1}\big), \qquad k = 1, \dots, K,
  \label{eq:loop}
\end{equation}
where $F$ applies frame attention within each view and then global attention across all views~\cite{wang2025vggt}, and decoder heads read $z_K$ to predict each view's camera, ray map and depth. The pose and intrinsics of each view are recovered from its ray map~\cite{lin2025depthanything3}, and lifting the depth with them gives the pointmap. The step times $t_1, \dots, t_K$ are evenly spaced from 0 to 1, with $t_{K+1} = t_K = 1$ on the last step, so within the trained range $K$ only sets how finely the same refinement is divided and can be chosen at inference; we use DVLT's default $K = 16$.

Because the weights are shared, the block knows where it is in the recursion only through a depth-scaling gate. A small MLP maps a sinusoidal embedding $\gamma$ of the current interval to three per-channel scales,
\begin{equation}
  s_{\mathrm{attn}}, s_{\mathrm{mlp}}, s_{\mathrm{out}} = 1 + \mathrm{MLP}\big(\gamma(t_k, t_{k+1})\big) \in \mathbb{R}^{C}.
  \label{eq:base-gate}
\end{equation}
The frame and the global attention block each have such a gate, and within a block that maps tokens $z$ to $\hat{z}$ it scales the residual update,
\begin{align}
  z' &= z + s_{\mathrm{attn}} \odot \mathrm{LS}_1\big(\mathrm{Attn}(\mathrm{LN}_1(z))\big), \\
  z'' &= z' + s_{\mathrm{mlp}} \odot \mathrm{LS}_2\big(\mathrm{MLP}(\mathrm{LN}_2(z'))\big), \\
  \hat{z} &= s_{\mathrm{out}} \odot z'' ,
\end{align}
where $\mathrm{LN}$ is layer normalisation, $\mathrm{LS}$ is LayerScale~\cite{touvron2021cait} and $\odot$ is channel-wise multiplication. $s_{\mathrm{attn}}$ and $s_{\mathrm{mlp}}$ set how much of the current update is written into a token, and $s_{\mathrm{out}}$ how much of its accumulated state carries into the next step. The gates vary channelwise and are broadcast across patches.

The event student is a copy of this model that differs in three places: an input stem for events (Sec.~\ref{sec:method-tokens}), LoRA~\cite{hu2022lora} adapters in all twelve DINOv2 blocks and in the shared block, and the gating of Sec.~\ref{sec:method-patchwise}. With about 1\% of the parameters trained and the decoders frozen, the student inherits the teacher's geometric knowledge and has only to reach the same final tokens from a different sensor.

\subsection{Feature distillation}
\label{sec:method-distill}

We distil the teacher into the student on sets of $S$ consecutive views, each consisting of an RGB frame and the events of its time window. In EventScape, on which we distil, the frame and the event camera share a viewpoint and a pixel grid, so the $n$-th token of the student covers the same part of the scene as the $n$-th token of the teacher. The student is moreover initialised from the teacher and adapted only by LoRA, so it starts inside the teacher's token space; the two can therefore be compared directly in latent space (Sec.~\ref{sec:method-loss}).

\subsubsection{Event tokenisation}
\label{sec:method-tokens}

For each frame, the events in its time window are accumulated into a voxel grid of five time bins, in which every event adds its polarity, $+1$ for brightening and $-1$ for darkening, to its pixel, split linearly between the two nearest bins~\cite{zhu2019unsupervised}. Each bin is then divided by its Euclidean norm over the image, which makes the grid invariant to how many events the window happens to contain. The input stem turns this grid into a three-channel image that DINOv2 can read: at each pixel the three channels hold the accumulated positive polarity, the accumulated absolute polarity and the accumulated negative polarity, summed over the bins, and each channel is divided by its maximum over the frame. This image is normalised with a per-channel scale and shift that start at the ImageNet normalisation values and are learned, so that they can adapt to events.

\subsubsection{Distillation loss}
\label{sec:method-loss}

We train the student to reproduce the teacher's final tokens $z_K$ rather than its predictions. All three heads read $z_K$ and are frozen, so a student whose tokens match the teacher's obtains cameras, rays and depth from the same decoders, and a single loss covers all three outputs.

Because DVLT is trained with $K$ drawn from $\mathrm{Beta}(2,1)$ rescaled to $[8,16]$~\cite{burzio2026dejaview}, its outputs are usable at any $K$ in that range, and we therefore treat the teacher's $z_K$ as a target at each such $K$. We draw one $K$ from this distribution per iteration, run both models for $K$ steps and match their final tokens of every view, including the camera and register tokens,
\begin{equation}
  \mathcal{L}_{\mathrm{distill}} = \mathbb{E}_{K}\big[\, d(z_K^{\mathrm{s}}, z_K^{\mathrm{t}}) \big],
  \label{eq:distill}
\end{equation}
allowing the student to run with any such $K$ at inference. The distance combines a cosine term, which aligns the direction of each token, with a magnitude-weighted $\ell_1$ term,
\begin{equation}
  \begin{gathered}
    d(z^{\mathrm{s}}, z^{\mathrm{t}}) = \mathbb{E}_n\Big[\tfrac{w_n}{C}\big\lVert z^{\mathrm{s}}_n - z^{\mathrm{t}}_n\big\rVert_1\Big] + 1 - \mathbb{E}_n\big[\cos(z^{\mathrm{s}}_n, z^{\mathrm{t}}_n)\big], \\
    w_n = \frac{\lVert z^{\mathrm{t}}_n \rVert_2}{\max_m \lVert z^{\mathrm{t}}_m \rVert_2},
  \end{gathered}
  \label{eq:token-distance}
\end{equation}
where $n$ runs over all tokens and $w_n$, the norm of the teacher's token relative to the largest in its frame, is treated as a constant. We match magnitude as well as the direction because the frozen decoders read the tokens as they are~\cite{ranzinger2024amradio}.

\subsection{Sparsity-aware adaptive gating}
\label{sec:method-patchwise}

During distillation the gates of Eq.~\eqref{eq:base-gate} keep the values learned on RGB and, as they depend only on the step, apply the same scales to every patch. This suits event images poorly: a few patches are dense with events while large regions are nearly empty (Sec.~\ref{sec:intro}), so the scale that suits one patch at a given step may not suit another.

\begin{figure}[t]
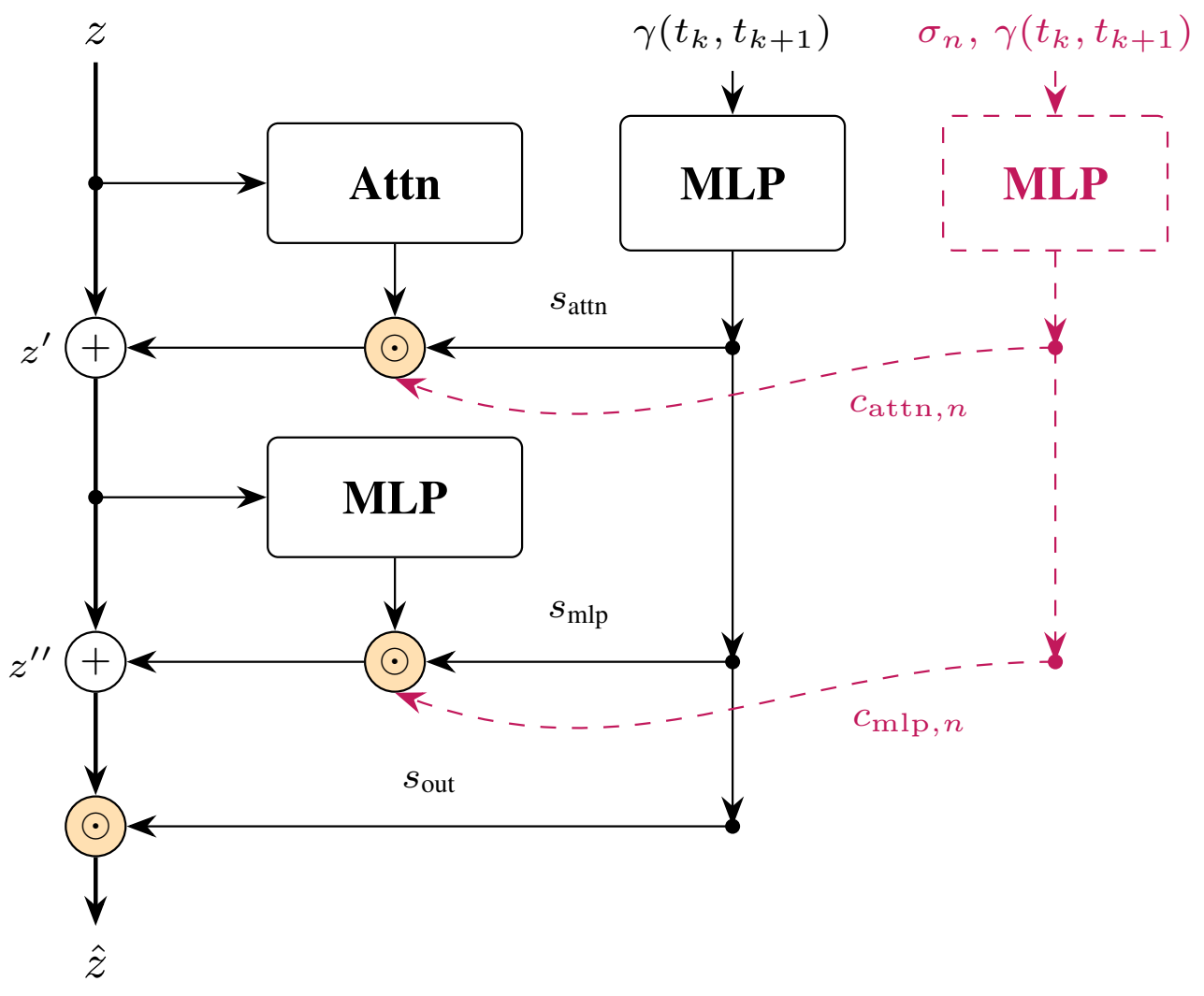

  \centering
  \includestandalone[width=\linewidth]{figures/standalone-diagram}
  \caption{Sparsity-aware adaptive gating in one attention block, applied in
  both frame and global attention. The step gate (centre) scales the attention
  update, the MLP update and the output token, and is fine-tuned for events.
  A patchwise correction (dashed), a small MLP over the event density $\sigma$
  and the interval embedding, rescales the attention and MLP gates of each
  patch.}
  \label{fig:patchwise-heads}
\end{figure}

Sparsity-aware adaptive gating (SAAG) adapts the gates to events in two ways. First, it fine-tunes the gate MLPs of both attention blocks together with the adapters, recalibrating the step schedule for events. Second, it makes the gates depend on the event density $\sigma_n \in [0,1]$ of each patch $n$, the average absolute polarity in the patch relative to the most active patch in the frame, computed once from the input and fixed across steps. In both the frame and the global attention block, a small MLP maps this density and the interval embedding to a correction of each gate (Fig.~\ref{fig:patchwise-heads}),
\begin{equation}
  \begin{aligned}
    (c_{\mathrm{attn},n}, c_{\mathrm{mlp},n}) &= \mathrm{MLP}\big(\sigma_n, \gamma(t_k, t_{k+1})\big), \\
    s'_{g,n} &= s_g \big(1 + c_{g,n}\big),
  \end{aligned}
  \label{eq:patchwise-gate}
\end{equation}
and, for each gate $g \in \{\mathrm{attn}, \mathrm{mlp}\}$, $s'_{g,n}$ replaces $s_g$ in the residual update of patch $n$. Because the correction multiplies the gate, it can strengthen or weaken the update of a patch. Each correction is a single value for the whole patch, applied equally to all channels, as it is driven by a per-patch density. Each MLP has one hidden layer and 2{,}114 parameters, and is trained together with the gates and adapters through $\mathcal{L}_{\mathrm{distill}}$.

%% file: figures/standalone-diagram.tex
\begin{tikzpicture}[>=Stealth, x=1mm, y=1mm]

\node[font=\small] (zkm1) at (0,0) {$z$};
\coordinate (split1) at ([yshift=-10mm]zkm1);

\node[block, minimum width=17mm] (attnblk) at ([xshift=20mm]split1) {\textbf{Attn}};
\node[gate] (gattn) at ([yshift=-11mm]attnblk) {$\odot$};
\node[add] (add1) at (gattn -| zkm1) {$+$};
\node[font=\scriptsize, anchor=east] at ([xshift=-1.5mm]add1) {$z'$};

\draw[resline] (zkm1) -- (add1);
\draw[branchline] (split1) -- (attnblk.west);
\draw[branchline] (attnblk) -- (gattn);
\draw[branchline] (gattn.west) -- (add1.east);
\fill (split1) circle (0.5mm);

\coordinate (split2) at ([yshift=-10mm]add1);
\node[block, minimum width=17mm] (mlpblk) at ([xshift=20mm]split2) {\textbf{MLP}};
\node[gate] (gmlp) at ([yshift=-11mm]mlpblk) {$\odot$};
\node[add] (add2) at (gmlp -| zkm1) {$+$};
\node[font=\scriptsize, anchor=east] at ([xshift=-1.5mm]add2) {$z''$};

\draw[resline] (add1) -- (add2);
\draw[branchline] (split2) -- (mlpblk.west);
\draw[branchline] (mlpblk) -- (gmlp);
\draw[branchline] (gmlp.west) -- (add2.east);
\fill (split2) circle (0.5mm);

\node[gate] (sout) at ([yshift=-11mm]add2) {$\odot$};
\node[font=\small] (zk) at ([yshift=-9mm]sout) {$\hat{z}$};
\draw[resline] (add2) -- (sout);
\draw[resline] (sout) -- (zk);

\coordinate (dscol) at ([xshift=14mm]attnblk.east);
\node[font=\scriptsize] (tpair) at (zkm1 -| dscol) {$\gamma(t_k,t_{k+1})$};
\node[block, minimum width=15mm, minimum height=9mm] (dmlp) at ([yshift=-10mm]tpair) {\textbf{MLP}};
\draw[branchline] (tpair) -- (dmlp.north);

\coordinate (tapA) at (gattn -| dscol);
\coordinate (tapB) at (gmlp -| dscol);
\coordinate (tapC) at (sout -| dscol);
\draw[branchline] (dmlp.south) -- (tapA);
\draw[branchline] (tapA) -- (tapB);
\draw[branchline] (tapB) -- (tapC);
\fill (tapA) circle (0.5mm);
\fill (tapB) circle (0.5mm);
\fill (tapC) circle (0.5mm);

\draw[branchline] (tapA) -- (gattn.east);
\draw[branchline] (tapB) -- (gmlp.east);
\draw[branchline] (tapC) -- (sout.east);
\node[font=\scriptsize, anchor=south, yshift=1mm] at ($(tapA)!0.5!(gattn.east)$) {$s_{\text{attn}}$};
\node[font=\scriptsize, anchor=south, yshift=1mm] at ($(tapB)!0.5!(gmlp.east)$) {$s_{\text{mlp}}$};
\node[font=\scriptsize, anchor=south, yshift=1mm] at ($(tapC)!0.5!(sout.east)$) {$s_{\text{out}}$};

\coordinate (ccol) at ([xshift=14mm]dmlp.east);
\node[font=\scriptsize, text=addborder] (cpair) at (zkm1 -| ccol) {$\sigma_n,\,\gamma(t_k,t_{k+1})$};
\node[block, draw=addborder, dashed, minimum width=15mm, minimum height=9mm] (cmlp) at ([yshift=-10mm]cpair)
  {\textbf{\textcolor{addborder}{MLP}}};
\draw[draw=addborder, dashed, branchline] (cpair) -- (cmlp.north);

\coordinate (ctapA) at (gattn -| ccol);
\coordinate (ctapB) at (gmlp -| ccol);
\draw[draw=addborder, dashed, branchline] (cmlp.south) -- (ctapA);
\draw[draw=addborder, dashed, branchline] (ctapA) -- (ctapB);
\fill[addborder] (ctapA) circle (0.5mm);
\fill[addborder] (ctapB) circle (0.5mm);
\draw[draw=addborder, dashed, branchline] (ctapA) to[out=180,in=-20] (gattn.south);
\draw[draw=addborder, dashed, branchline] (ctapB) to[out=180,in=-20] (gmlp.south);
\node[font=\scriptsize, text=addborder, anchor=north] at ($(ctapA)!0.22!(gattn.south)+(0,-1.5mm)$) {$c_{\mathrm{attn},n}$};
\node[font=\scriptsize, text=addborder, anchor=north] at ($(ctapB)!0.22!(gmlp.south)+(0,-1.5mm)$) {$c_{\mathrm{mlp},n}$};

\end{tikzpicture}

%% file: sections/experiments.tex
\section{Experiments}
\label{sec:experiments}

\subsection{Teacher and baselines}
\label{sec:experiments-baselines}

The teacher is the frozen DVLT model~\cite{burzio2026dejaview}. It sees RGB frames rather than events, so we report it in italics as a reference and not as a competitor. Our event student, Déjà-Events, is distilled on the EventScape training split without any depth or pose ground truth, for 20{,}000 steps on clips of 2 to 18 views, with LoRA rank 16 and 1.6\,M trained parameters, using AdamW with a cosine schedule from peak learning rate $3\times10^{-4}$ (a tenth of that for adapters on pretrained weights) and weight decay 0.05.

We compare against three kinds of event-based methods, which also form the groups of Table~\ref{tab:baselines-pointmap}. The depth-only networks are E2Depth~\cite{hidalgocarrio2020e2depth} and EReFormer~\cite{liu2022ereformer}, retrained on EventScape, and DepthAnyEvent and its recurrent variant DepthAnyEvent-R~\cite{bartolomei2025depthanyevent}, abbreviated DAE and DAE-R, which build on Depth Anything V2~\cite{yang2024depthanythingv2} and whose released checkpoints were trained on EventScape. EventVGGT~\cite{ren2026eventvggt} predicts depth and pose with two separately trained checkpoints: \emph{Option~A} takes depth from the depth checkpoint and pose and intrinsics from the pose checkpoint, and \emph{Option~B} takes all three from the pose checkpoint, making it a single model. We run both released checkpoints with the authors' inference code and event representation, unmodified. For visual odometry we use DEVO~\cite{klenk2024devo}, and pairing its poses with each depth network gives \emph{modular} pipelines, which are lifted to 3D with the ground-truth intrinsics, whereas Déjà-Events, its teacher and EventVGGT use the intrinsics they predict.

\subsection{Datasets}
\label{sec:experiments-datasets}

EventScape~\cite{gehrig2021ramnet} is simulated in CARLA~\cite{dosovitskiy2017carla} with a $512{\times}256$ event camera and dense depth and poses; we distil on Town01 to Town03 and test on the 116 sequences of its released Town05 test set, capping depth at 200\,m. MVSEC~\cite{zhu2018mvsec} records driving with a $346{\times}260$ DAVIS346, and we use four of its outdoor drives, one by day and three at night, with the lidar depth capped at 80\,m. DSEC~\cite{gehrig2021dsec} records driving with a $640{\times}480$ event camera; since its test split has no public ground truth, we follow~\cite{nam2022secff,bartolomei2025depthanyevent} and use ten validation sequences held out from its training split, with lidar-based depth that DSEC limits to 50\,m and pseudo ground-truth poses that we obtain by lidar odometry~\cite{vizzo2023kissicp}.

\subsection{Evaluation protocol}
\label{sec:experiments-protocol}

Each test sequence is cut into non-overlapping 18-frame windows, which every method processes at its own input resolution; EReFormer's positional embeddings are hardcoded for a $224{\times}224$ input, so on DSEC its predictions and the ground truth are centre-cropped to 16\% of the frame, and its numbers there are not strictly comparable with the rest. Because DAE and DAE-R predict depth only up to scale and shift, depth is aligned with a median-based scale and shift. Pose and pointmaps are aligned with Sim(3), and ground-truth points are expressed in the window's first camera. None of the baselines were trained on MVSEC or DSEC.

Depth is measured by the absolute relative error (AbsRel) and $\Delta_1$, the fraction of pixels within a factor of 1.25 of ground truth. Pose is measured by the absolute trajectory error (ATE) in meters and AUC$_{20}$, the area under the accuracy curve, up to 20$^\circ$, of the larger of each frame pair's rotation and translation-direction errors~\cite{wang2023posediffusion}. For pointmaps, we follow DVLT~\cite{burzio2026dejaview} and report AbsRel, the point error relative to the ground-truth distance, and Inl@25, the fraction of points within 25\% relative error, which is the 3D analogue of $\Delta_1$.

%% file: sections/results.tex
\section{Results}
\label{sec:results}

\subsection{Depth}
\label{sec:results-depth}
\label{sec:results-baselines}

On EventScape, Déjà-Events is ahead of every event-based method on both depth metrics (Table~\ref{tab:baselines-depth}). Its closest competitor is EventVGGT, whose AbsRel is 13\% higher. On MVSEC, EventVGGT is the most accurate event-based method, and Déjà-Events is second.

\begin{figure*}[t]
  \centering
  \includegraphics[width=\textwidth]{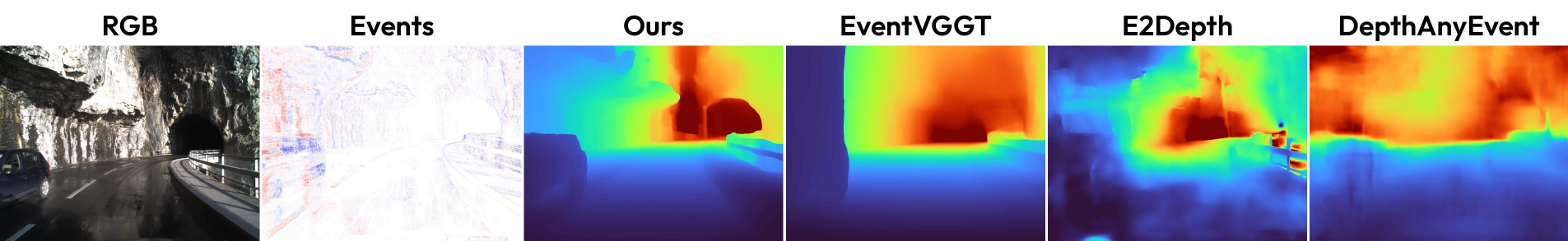}
  \caption{\textbf{Zero-shot depth on DSEC.} Frame 685 of
\texttt{interlaken\_00\_b}, from a 50\,ms event window. All columns show the
identical $476\times630$ crop of DSEC's event camera. This is a DSEC test-split
sequence, which ships no public disparity, so the comparison is qualitative;
no model was trained on DSEC and all baseline checkpoints are EventScape-trained.
Each depth map is colourised against its own 1st/99th percentile.}
  \label{fig:dsec-depth}
\end{figure*}

On DSEC the student leads every event-based method on both metrics, with about half the relative error of EventVGGT. The RGB teacher keeps a large lead on this dataset, since DSEC's frame camera stays well exposed even in its darkest sequences. Fig.~\ref{fig:dsec-depth} shows depth maps from DSEC.

\begin{table}[t]
  \centering
  \footnotesize
  \caption{Depth estimation.}
  \label{tab:baselines-depth}
  \setlength{\tabcolsep}{2pt}
  \begin{tabular}{lcccccc}
    \toprule
    & \multicolumn{2}{c}{EventScape} & \multicolumn{2}{c}{MVSEC} & \multicolumn{2}{c}{DSEC} \\
    \cmidrule(lr){2-3}\cmidrule(lr){4-5}\cmidrule(lr){6-7}
    Model & AbsRel$\downarrow$ & $\Delta_1$$\uparrow$ & AbsRel$\downarrow$ & $\Delta_1$$\uparrow$ & AbsRel$\downarrow$ & $\Delta_1$$\uparrow$ \\
    \midrule
    \textit{DVLT teacher} & \textit{0.111} & \textit{0.898} & \textit{0.290} & \textit{0.556} & \textit{0.060} & \textit{0.965} \\
    Déjà-Events & \textbf{0.144} & \textbf{0.843} & \underline{0.298} & \underline{0.608} & \textbf{0.145} & \textbf{0.841} \\
    EventVGGT & \underline{0.162} & \underline{0.751} & \textbf{0.275} & \textbf{0.618} & \underline{0.282} & \underline{0.550} \\
    E2Depth & 0.283 & 0.543 & 0.381 & 0.417 & 0.408 & 0.416 \\
    EReFormer & 0.241 & 0.648 & 0.411 & 0.421 & 0.409$^*$ & 0.511$^*$ \\
    DAE & 0.397 & 0.329 & 0.418 & 0.367 & 0.297 & 0.486 \\
    DAE-R & 0.640 & 0.201 & 0.544 & 0.279 & 0.400 & 0.372 \\
    \bottomrule
  \end{tabular}
  \tabnote{MVSEC and DSEC are zero-shot. \textbf{Bold}: best; \underline{underline}: second best among event-based methods. $^*$EReFormer's DSEC predictions are centre-cropped (Sec.~\ref{sec:experiments-protocol}), so they are not strictly comparable and are not ranked.}
\end{table}

\subsection{Pose}
\label{sec:results-pose}

On EventScape, Déjà-Events's trajectory error is a little over half that of DEVO and a small fraction of EventVGGT's (Table~\ref{tab:baselines-pose}). On the real sensors, DEVO has the lowest translation errors, with the student second. EventVGGT's camera head does not recover the ego-motion on any dataset: its translation directions are close to chance. The teacher remains more accurate than the student on EventScape and DSEC. MVSEC is the exception, where the student's trajectory error falls below the teacher's owing to the night drives, in which low illumination degrades the RGB input (Sec.~\ref{sec:results-distill}).

\begin{table}[t]
  \centering
  \footnotesize
  \caption{Camera pose.}
  \label{tab:baselines-pose}
  \setlength{\tabcolsep}{2pt}
  \begin{tabular}{lcccccc}
    \toprule
    & \multicolumn{2}{c}{EventScape} & \multicolumn{2}{c}{MVSEC} & \multicolumn{2}{c}{DSEC} \\
    \cmidrule(lr){2-3}\cmidrule(lr){4-5}\cmidrule(lr){6-7}
    Model & ATE$\downarrow$ & AUC$_{20}$$\uparrow$ & ATE$\downarrow$ & AUC$_{20}$$\uparrow$ & ATE$\downarrow$ & AUC$_{20}$$\uparrow$ \\
    \midrule
    \textit{DVLT teacher} & \textit{0.013} & \textit{0.953} & \textit{0.144} & \textit{0.732} & \textit{0.043} & \textit{0.897} \\
    Déjà-Events & \textbf{0.036} & \textbf{0.887} & \underline{0.095} & \underline{0.717} & \underline{0.105} & \underline{0.811} \\
    EventVGGT & 0.796 & 0.029 & 1.327 & 0.031 & 3.785 & 0.018 \\
    DEVO & \underline{0.062} & \underline{0.850} & \textbf{0.034} & \textbf{0.731} & \textbf{0.027} & \textbf{0.972} \\
    \bottomrule
  \end{tabular}
  \tabnote{EventVGGT uses its released pose checkpoint, for which no quantitative pose results have been reported~\cite{ren2026eventvggt}; on the authors' released example, our inference reproduces their predicted camera.}
\end{table}

\subsection{Pointmap}
\label{sec:results-pointmap}

\begin{figure}[t]
  \centering
  \includegraphics[width=0.5\linewidth]{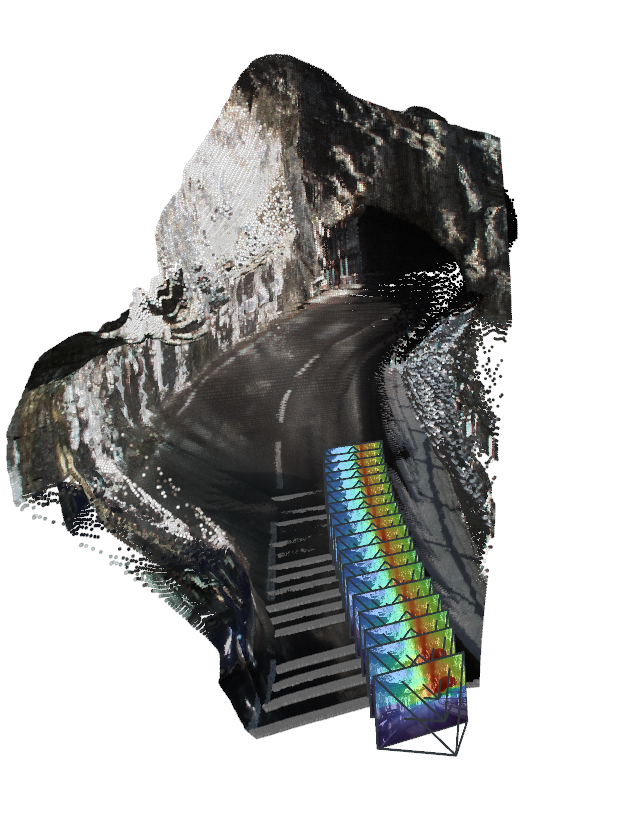}
  \caption{Pointmap from the 18-view window of \texttt{interlaken\_00\_b}
  containing the frame of Fig.~\ref{fig:dsec-depth}, predicted by our student
  from events alone; points are coloured with the aligned RGB frames for
  legibility, and the frustums are the student's predicted cameras.}
  \label{fig:dsec-pointmap}
\end{figure}

On EventScape, Déjà-Events has the lowest pointmap error of all event-based methods (Table~\ref{tab:baselines-pointmap}), 43\% below EventVGGT Option~A and within 8\% of its teacher, while the modular pipelines trail further despite DEVO's poses and the true intrinsics. The ordering holds zero-shot on the real sensors. On MVSEC, Déjà-Events leads every event-based method and also reconstructs the scene more accurately than its teacher, although the teacher has the lower depth error, and on DSEC it leads with about half the relative error of Option~A. Fig.~\ref{fig:dsec-pointmap} shows a full window reconstructed on DSEC, together with the cameras the student predicts for it.

Two effects explain these rankings. First, pose errors matter less than depth errors. A window covers only a short stretch of the drive, so a wrong pose shifts the points by at most the distance travelled, which is small compared with their depth. This is why EventVGGT still reconstructs reasonably well despite its near-static cameras (Table~\ref{tab:baselines-pose}): its two options use the same poses and intrinsics, and Option~A, which has the better depth, is ahead of Option~B on every dataset. Second, the intrinsics matter when their error is large. On MVSEC, the teacher predicts a focal length 2.8 times too long and Déjà-Events one 1.6 times too long, which is why the teacher reconstructs the scene worse than its student despite better depth. Smaller errors cost little: EventVGGT's focal length is about a quarter too short on every dataset, and that of Déjà-Events is 30\% too short on DSEC, where it still leads.

\begin{table}[t]
  \centering
  \footnotesize
  \caption{Pointmap reconstruction.}
  \label{tab:baselines-pointmap}
  \label{tab:mvsec-pointmap-zeroshot}
  \setlength{\tabcolsep}{1.7pt}
  \begin{tabular}{lcccccc}
    \toprule
    & \multicolumn{2}{c}{EventScape} & \multicolumn{2}{c}{MVSEC} & \multicolumn{2}{c}{DSEC} \\
    \cmidrule(lr){2-3}\cmidrule(lr){4-5}\cmidrule(lr){6-7}
    Model & AbsRel$\downarrow$ & Inl@25$\uparrow$ & AbsRel$\downarrow$ & Inl@25$\uparrow$ & AbsRel$\downarrow$ & Inl@25$\uparrow$ \\
    \midrule
    \textit{DVLT teacher} & \textit{0.187} & \textit{0.794} & \textit{0.350} & \textit{0.452} & \textit{0.058} & \textit{0.976} \\
    Déjà-Events & \textbf{0.202} & \textbf{0.789} & \textbf{0.297} & \textbf{0.575} & \textbf{0.148} & \textbf{0.855} \\
    EventVGGT A & \underline{0.356} & \underline{0.451} & \underline{0.347} & \underline{0.514} & \underline{0.301} & \underline{0.577} \\
    EventVGGT B & 0.700 & 0.201 & 0.459 & 0.368 & 0.331 & 0.538 \\
    \addlinespace
    \multicolumn{7}{l}{\itshape modular (DEVO pose + GT intrinsics)} \\
    E2Depth & 0.548 & 0.256 & 0.382 & 0.440 & 0.326 & 0.564 \\
    EReFormer & 0.404 & 0.427 & 0.409 & 0.387 & 0.242$^*$ & 0.661$^*$ \\
    DAE & 0.610 & 0.266 & 0.520 & 0.269 & 0.336 & 0.412 \\
    DAE-R & 0.895 & 0.196 & 0.531 & 0.297 & 0.304 & 0.555 \\
    \bottomrule
  \end{tabular}
  \tabnote{MVSEC and DSEC are zero-shot. \textbf{Bold}: best; \underline{underline}: second best among event-based methods. $^*$EReFormer's DSEC predictions are centre-cropped (Sec.~\ref{sec:experiments-protocol}), so they are not strictly comparable and are not ranked.}
\end{table}

\subsection{Distillation on MVSEC day 1}
\label{sec:results-distill}

Table~\ref{tab:mvsec-daynight} reports results on MVSEC night1, where low illumination degrades the teacher's RGB input. It also adds Déjà-Events distilled for one more epoch on a daytime drive, which leaves night1 held out, with the same feature distillation and no ground truth.

At night, Déjà-Events already has less than half its teacher's trajectory error and a more accurate reconstruction, while the teacher keeps the lower relative depth error. The distillation on day 1 transfers to night: relative depth error falls below the teacher's and pose accuracy reaches the teacher's level, while the trajectory error is unchanged. The reconstruction, however, becomes less accurate than the zero-shot model's. We attribute this to the camera intrinsics through which the pointmap is formed: the extra epoch pulls the predicted focal length towards the teacher's, 2.8 times the true value, whereas the zero-shot model overestimates it by only 1.6 times.

\begin{table}[t]
  \centering
  \footnotesize
  \caption{MVSEC outdoor\_night1.}
  \label{tab:mvsec-daynight}
  \label{tab:distill-day1}
  \setlength{\tabcolsep}{2pt}
  \begin{tabular}{lccccc}
    \toprule
    & \multicolumn{2}{c}{Depth} & \multicolumn{2}{c}{Pose} & Pointmap \\
    \cmidrule(lr){2-3}\cmidrule(lr){4-5}\cmidrule(lr){6-6}
    Model & AbsRel$\downarrow$ & $\Delta_1$$\uparrow$ & ATE$\downarrow$ & AUC$_{20}$$\uparrow$ & AbsRel$\downarrow$ \\
    \midrule
    \textit{DVLT teacher} & \textit{0.339} & \textit{0.507} & \textit{0.156} & \textit{0.823} & \textit{0.424} \\
    Déjà-Events & 0.373 & 0.565 & \textbf{0.065} & 0.790 & \textbf{0.347} \\
    Déjà-Events + day1 & \textbf{0.324} & \textbf{0.567} & 0.066 & \textbf{0.821} & 0.418 \\
    \bottomrule
  \end{tabular}
\end{table}

\subsection{Compute cost}
\label{sec:results-compute}

Table~\ref{tab:compute} compares model size, compute and memory on MVSEC. Distillation trains only about 1\% of the student's parameters, and the event student needs about 1\% more compute and 2\% more memory than its RGB teacher. EventVGGT has about ten times as many parameters, two to almost four times the compute depending on the configuration, and about 2.5 times the memory. The modular pipelines use about the same compute as the student and half its memory, but they train 9 to 58 times as many parameters on events. 

\begin{table}[h]
  \centering
  \footnotesize
  \setlength{\tabcolsep}{3pt}
  \caption{Model size and compute.}
  \label{tab:compute}
  \begin{tabular}{lrrrr}
    \toprule
    & & & GFLOPs & Peak mem. \\
    Model & Params & Trained & / frame $\downarrow$ & (GiB) $\downarrow$ \\
    \midrule
    \textit{DVLT teacher} & \textit{117.1} & \textit{--} & \textit{597.7} & \textit{1.62} \\
    Déjà-Events & 118.4 & 1.60 & 604.6 & 1.65 \\
    \midrule
    EventVGGT A & 1161.1$^\ddagger$ & 3.15$^\dagger$ & 2209.0 & 4.21$^\ddagger$ \\
    EventVGGT B & 1159.5 & \textbf{1.57}$^\dagger$ & 1265.0 & 4.20 \\
    \midrule
    DEVO + E2Depth & 14.1 & 14.1 & 628.7 & 0.79 \\
    DEVO + EReFormer & 92.9 & 92.9 & 580.8 & 0.89 \\
    DEVO + DAE & 28.2 & 28.2 & \textbf{574.0} & \textbf{0.74} \\
    DEVO + DAE-R & 29.5 & 29.5 & 580.0 & 0.75 \\
    \bottomrule
  \end{tabular}
  \tabnote{Params and Trained (parameters trained on events) in millions; peak memory for one inference pass on an A40. $^\dagger$LoRA parameters. $^\ddagger$One shared backbone with different LoRA adapters.}
\end{table}

\subsection{Ablations}
\label{sec:results-ablations}

\subsubsection{Sparsity-aware adaptive gating}
\label{sec:results-patchwise}

\begin{figure}[t]
  \centering
  \includegraphics[width=\linewidth]{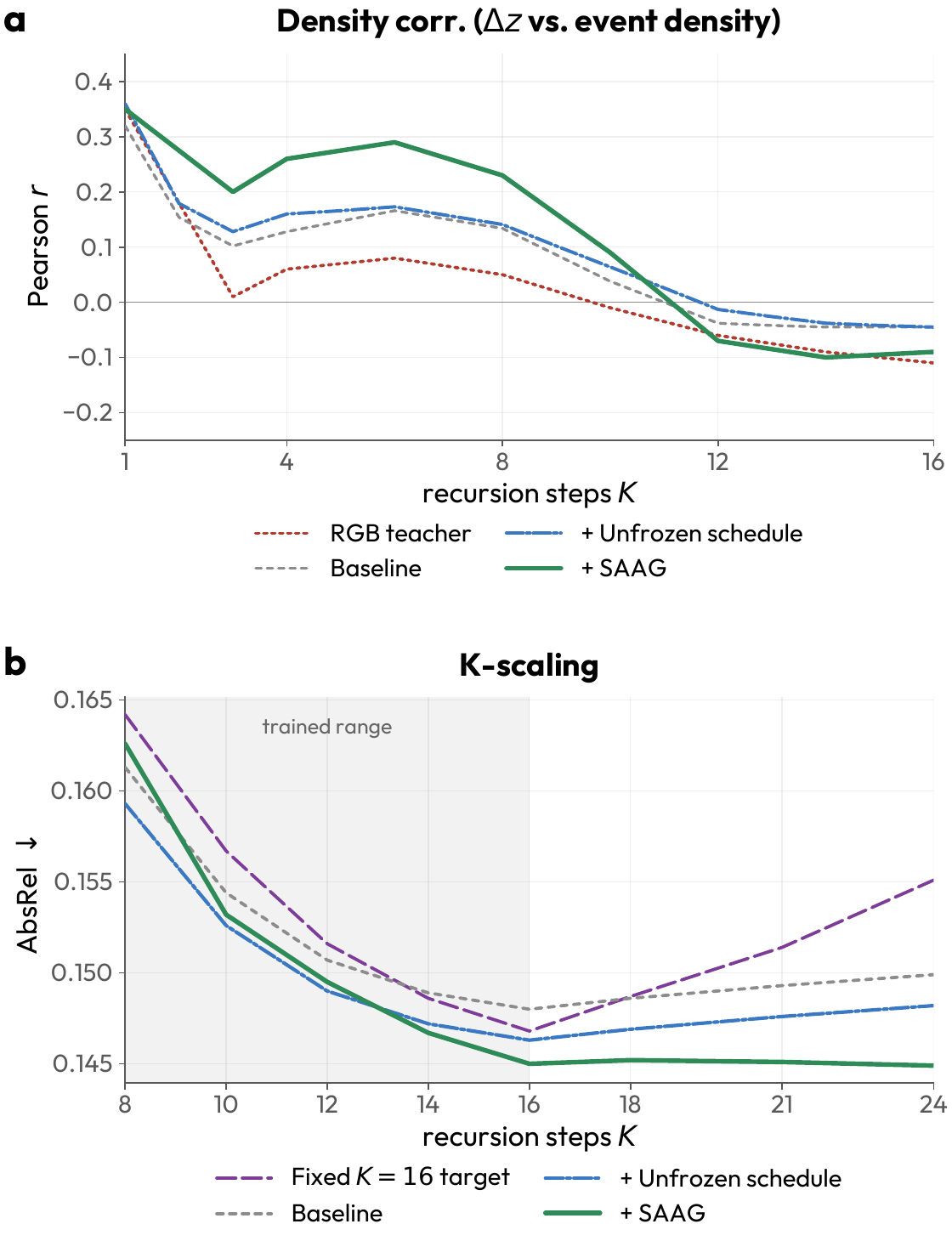}
  \caption{\textbf{(a)} Per-step correlation between the token update
  $\Delta z_k$ and event density, for the frozen-schedule student, the same
  student with an unfrozen schedule, our final patchwise correction, and an
  RGB teacher probed on the same clips as a control. \textbf{(b)} AbsRel on
  DSEC (zero-shot) as inference steps $K$ are swept past the $[8,16]$ range
  seen in training.}
  \label{fig:patchwise-mechanism}
\end{figure}

The payoff of conditioning on density is at inference time.
In Fig.~\ref{fig:patchwise-mechanism}b, a student trained against a fixed
$K{=}16$ target degrades as soon as $K$ moves past it, while sampled-$K$
training (Sec.~\ref{sec:method-distill}, used by every other curve here)
extrapolates smoothly, and unfreezing the schedule shifts accuracy down
fairly uniformly without changing that shape. On DSEC, ours is the only arm that does not degrade past that range, which
leaves recursion depth a usable compute knob rather than a fixed operating
point.

Fig.~\ref{fig:patchwise-mechanism}a shows the mechanism. Every arm, the RGB
teacher control included, runs from a positive density--update correlation to
a negative one, so the crossover is generic; what SAAG changes is its
strength, roughly doubling the schedules' correlation over the first two
thirds of the recursion while the teacher's is half theirs.

\begin{table}[h]
  \centering
  \footnotesize
  \caption{Step-scale ablation.}
  \label{tab:patchwise}
  \begin{tabular}{lccc}
    \toprule
    Student configuration & AbsRel $\downarrow$ & $\Delta_1$ $\uparrow$ & ATE $\downarrow$ \\
    \midrule
    \textit{DVLT teacher (RGB, ceiling)} & \textit{0.111} & \textit{0.898} & \textit{0.013} \\
    \midrule
    Distillation only, frozen schedule & 0.147 & 0.839 & 0.036 \\
    \quad + unfrozen schedule & 0.146 & 0.841 & 0.036 \\
    \quad + SAAG (ours) & \textbf{0.144} & \textbf{0.843} & 0.036 \\
    \bottomrule
  \end{tabular}
  \tabnote{EventScape test split. The teacher is a ceiling, not a comparable arm.}
\end{table}

Table~\ref{tab:patchwise} adds the two step-scale changes of
Sec.~\ref{sec:method-patchwise} cumulatively on top of sampled-$K$ training:
at the default $K{=}16$ they lower AbsRel by 0.003 and leave trajectory error
unchanged. The schedule matters more against a conventionally distilled
student: a fixed-$K{=}16$ target is competitive only at the step count it was
trained for and degrades beyond it, whereas ours holds out to $K{=}24$
(Fig.~\ref{fig:patchwise-mechanism}b). Everything else is inherited from DVLT
unchanged -- the recursion, the gate of Eq.~\eqref{eq:base-gate}, the
decoders, the $K$ distribution and the optimiser settings -- so the arms
differ only in the component under test.

%% file: sections/discussion.tex
\section{Discussion}
\label{sec:discussion}

\paragraph*{Accurate and cheap Event 3D reconstruction} Déjà-Events gives the most accurate event-only reconstruction we measure on all three datasets, and it does so from a model an order of magnitude smaller than its closest competitor: 118\,M parameters against EventVGGT's 1.16\,B, at less than half its compute and 40\% of its memory (Table~\ref{tab:compute}). Only 1.6\,M of those parameters are trained, and none of that training uses depth or pose ground truth, because the geometric prior and the decoders are inherited from the RGB teacher and only the encoding of events has to be learned.
\paragraph*{Bounded by the teacher's input} Outside MVSEC the student stays behind its teacher by a margin related to the quality of the frames the teacher receives rather than to the distillation. It is widest on DSEC, whose colour frames stay well exposed even in its darkest sequences, and smallest on MVSEC, whose low-resolution greyscale frames come from three night drives and one daylight drive; there the student already reconstructs the scene more accurately than its teacher, and distillation on the day sequence brings its night depth error below the teacher's as well (Table~\ref{tab:mvsec-daynight}). Because the student is trained to reproduce its teacher, it can overtake it only where the teacher's input fails, and driving data rarely leaves the regime its frame cameras handle. Future work should therefore evaluate on agile flight, where motion blur degrades the frames, to test whether events overtake them there.

\paragraph*{Bounded by the teacher's camera} How much the predicted camera costs can be measured by projecting the same depth and poses through the ground-truth calibration. On MVSEC this lowers the teacher's pointmap error by 18\% and the student's by 7\%, leaving the two within 0.013 of each other, and it turns the student distilled on day 1 from the least into the most accurate reconstruction we obtain on that sensor, which confirms that its regression is one of intrinsics rather than of depth. On DSEC the same substitution removes about a sixth of the student's error and leaves the teacher well ahead, so the gap there is depth, not the camera. Intrinsics are nonetheless the cheapest of these errors to remove, because a deployed camera is calibrated: fitting the camera head of an already distilled student to that calibration would repair its weakest output while depth and pose remain label-free.

\paragraph*{Timing is the next axis} An event camera reports each brightness change as it happens, so events record both structure a blurred frame loses and when, inside the interval, that structure moved. The first is what the student already exploits: blur and saturation degrade the teacher's input while the event stream keeps its edges, which is why it overtakes its teacher where frames fail. The second it does not use. A frame integrates the interval into one image, so the teacher's tokens cannot represent when anything happened inside it, and our stem discards the same information by summing over the bins (Sec.~\ref{sec:method-tokens}), which leaves fast motion smeared across the summed image. Reading the bins would sharpen that and, more usefully, give geometry between frames rather than only at them. Doing so from events alone needs targets that vary within the interval, not a temporal encoder by itself.

%% file: sections/conclusion.tex
\section{Conclusion}
\label{sec:conclusion}

We presented Déjà-Events, an event-only feed-forward model that predicts
depth, camera pose and pointmaps in a single pass. It is distilled on paired
RGB and event recordings from a frozen RGB teacher, DVLT, which refines its
tokens by reusing one shared block, so it needs no depth or pose ground truth
and trains about 1\% of its parameters. Distilled only in simulation, it has
the lowest pointmap error of the event-based methods we evaluate on
EventScape and, zero-shot, on the real MVSEC and DSEC sensors, and the lowest
depth error on EventScape and DSEC, at about a tenth of the strongest baseline's
parameters.

Sparsity-aware adaptive gating conditions the shared block's gates on where events fire;
on DSEC it alone does not degrade as inference steps grow past the trained
range, leaving recursion depth a usable compute knob.

Events also record when the scene moved inside a frame interval; using this
to predict geometry between frames, not only at them, is the natural next
step.